\documentclass[fleqn,10pt]{wlscirep}
\usepackage[utf8]{inputenc}
\usepackage[T1]{fontenc}

\usepackage{txfonts}
\usepackage{url}
\usepackage{bm}
\usepackage{color}
\usepackage{enumitem}

\usepackage{algorithm}
\usepackage{algorithmic}

\usepackage[whole]{bxcjkjatype}

\title{Effectiveness of Binary Autoencoders for QUBO-Based Optimization Problems}

\author[1,*]{Tetsuro Abe}
\author[1]{Masashi Yamashita}
\author[1,2,3,4]{Shu Tanaka}

\affil[1]{Graduate School of Science and Technology, Keio University, Kanagawa 223-8522, Japan}
\affil[2]{Keio University Sustainable Quantum Artificial Intelligence Center (KSQAIC), Keio University, Tokyo 108-8345, Japan}
\affil[3]{Department of Applied Physics and Physico-Informatics, Keio University, Kanagawa 223-8522, Japan}
\affil[4]{Human Biology-Microbiome-Quantum Research Center (WPI-Bio2Q), Keio University, Tokyo 108-8345, Japan}

\affil[*]{Corresponding author : tetsu.abe1231@keio.jp}

\begin{abstract}
In black-box combinatorial optimization, objective evaluations are often expensive, so high quality solutions must be found under a limited budget. Factorization machine with quantum annealing (FMQA) builds a quadratic surrogate model from evaluated samples and optimizes it on an Ising machine.
However, FMQA requires binary decision variables, and for nonbinary structures such as integer permutations, the choice of binary encoding strongly affects search efficiency.
If the encoding fails to reflect the original neighborhood structure, small Hamming moves may not correspond to meaningful modifications in the original solution space, and constrained problems can yield many infeasible candidates that waste evaluations.
Recent work combines FMQA with a binary autoencoder (bAE) that learns a compact binary latent code from feasible solutions, yet the mechanism behind its performance gains is unclear.
Using a small traveling salesman problem as an interpretable testbed, we show that the bAE reconstructs feasible tours accurately and, compared with manually designed encodings at similar compression, better aligns tour distances with latent Hamming distances, yields smoother neighborhoods under small bit flips, and produces fewer local optima.
These geometric properties explain why bAE+FMQA improves the approximation ratio faster while maintaining feasibility throughout optimization, and they provide guidance for designing latent representations for black-box optimization.

\end{abstract}
\begin{document}

\flushbottom
\maketitle

\thispagestyle{empty}

\section*{Introduction}
Combinatorial optimization problems arise broadly across science and engineering, including transportation, structural design, and finance~\cite{yarkoni2022quantum, matsumori2022application, kanai2024annealing, takahashi2025effectiveness, nawa2023quantum}, yet many practically relevant instances are both computationally intractable and evaluation-intensive.
Many of these problems are NP-hard, and the size of the solution space grows exponentially with problem scale, making exhaustive search intractable. 
In addition, when the objective function is only accessible through simulations or experiments, each evaluation can be expensive, and the optimization task becomes a black-box problem in which the evaluation budget is a primary bottleneck.

A common strategy for expensive black-box optimization is to construct a surrogate model from evaluated data and to generate new candidate solutions by optimizing the surrogate model. 
Recently, approaches that represent the surrogate model as a quadratic unconstrained binary optimization (QUBO) enables the use of Ising machines, including quantum annealing (QA)~\cite{kadowaki1998quantum, das2008colloquium, tanaka2017quantum, tanahashi2019application, hauke2020perspectives, mohseni2022ising, chakrabarti2023quantum}, as efficient solvers for large discrete search spaces.
In this context, factorization machine with quantum annealing (FMQA)~\cite{kitai2020designing, inoue2022towards, nakano2026swift, tamura2025black} 
provides a practical framework by learning pairwise interactions via a factorization machine (FM)~\cite{rendle2010factorization, rendle2012factorization} to construct a quadratic surrogate model, and directly optimizing it on an Ising machine, thereby coupling surrogate model learning and combinatorial search in a unified binary formulation.

A key limitation of FMQA is that it assumes binary decision variables. 
To apply FMQA to problems naturally described by nonbinary structures, such as permutations, categorical variables, or multivalued discrete variables, one must introduce a binary encoding.
However, the choice of encoding can critically affect search efficiency: under an unsuitable encoding, a local move in Hamming space (e.g., a single-bit flip) may not preserve neighborhood relations or constraint structure in the original solution space~\cite{koshikawa2025efficient}.
As a result, neighborhood exploration becomes effectively random, learning of the surrogate model  can become unstable, and the optimization may waste evaluations. 
This issue is particularly severe in constrained problems, where feasible solutions can be extremely sparse in the binary space and many generated candidates violate constraints, further consuming the limited evaluation budget.

\begin{figure}[t]
    \centering
    \includegraphics[width=0.9\linewidth]{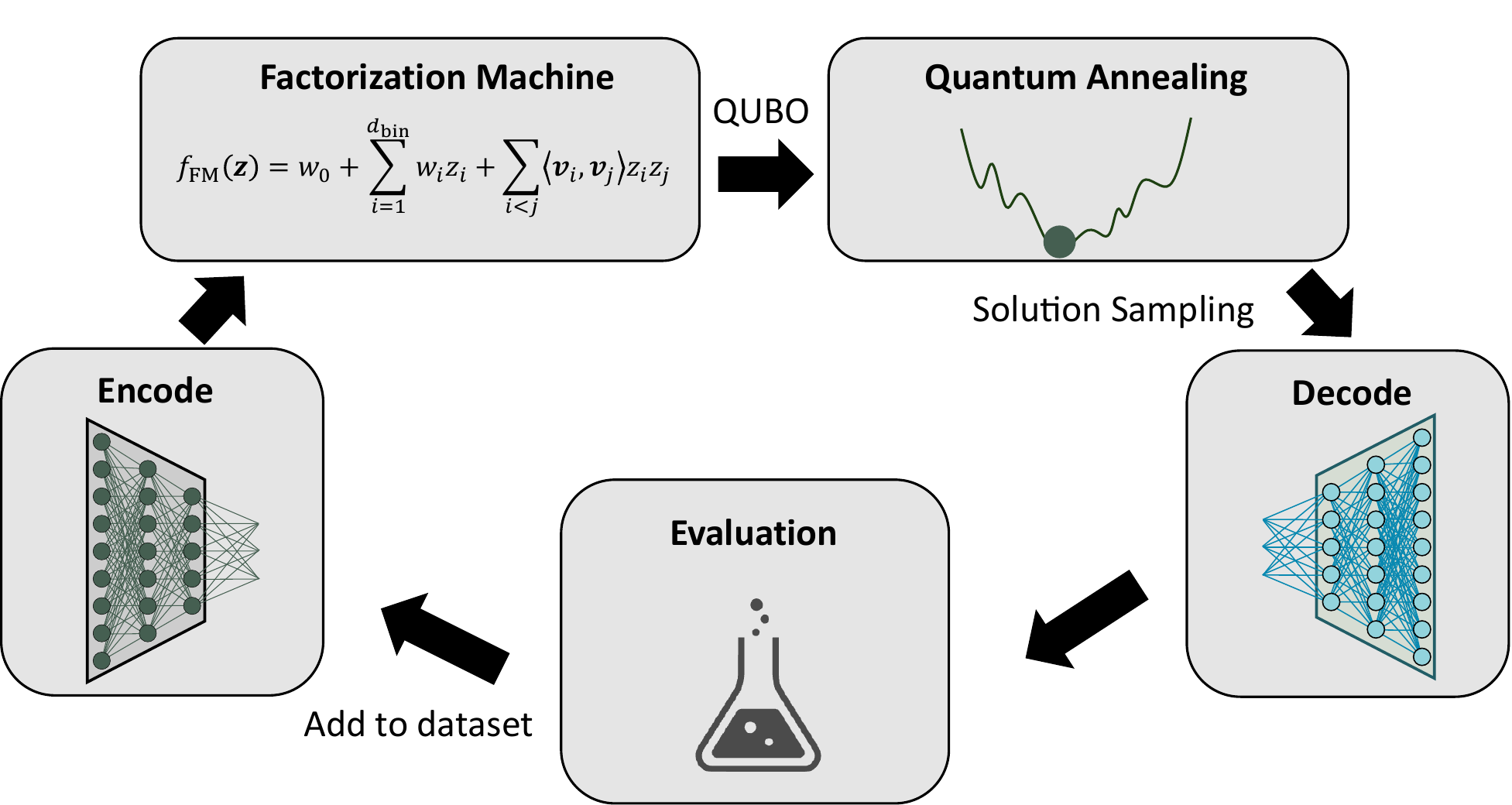}
    \caption{bAE+FMQA workflow for black-box optimization. A binary autoencoder (bAE) trained only on feasible solutions maps non-binary combinatorial solutions to a low-dimensional binary latent code. In the latent space, a factorization machine (FM) is trained on previously evaluated samples and their objective values, and its quadratic approximation is formulated as a QUBO and optimized using an Ising machine (e.g., quantum annealing). The optimized latent code is decoded back to the original solution representation and evaluated, and the resulting sample is added to the dataset for the next iteration, enabling black-box optimization without relying on handcrafted encodings.}
    \label{fig:bAE_FMQA}
\end{figure}

Recent studies have proposed using latent representations learned by generative models and performing search directly in the latent space~\cite{kingma2013auto,gomez2018automatic,kusner2017grammar,jin2018junction}.
In the context of QUBO-based optimization, such representations offer a promising way to alleviate the sensitivity to handcrafted binary encodings.
To mitigate these difficulties, the bAE+FMQA framework employs a binary autoencoder (bAE)~\cite{baynazarov2019binary} to map non-binary solutions into a compact binary latent code and runs FMQA over the latent space. 
This approach aims to reduce reliance on handcrafted encodings and instead learn a binary representation that is more suitable for search from problem-specific data. Empirical effectiveness has been reported in applications such as antimicrobial peptide and molecular design~\cite{tucs2023quantum, mao2023chemical}, yet the mechanisms that make bAE+FMQA work remain insufficiently understood (Fig.~\ref{fig:bAE_FMQA}). 
In particular, it is unclear which properties of the learned latent geometry enable efficient optimization. 
We systematically analyze three key aspects of the latent representation: (i) preservation of distance structure between solutions, (ii) locality induced by small bit flips in the binary space, and (iii) how feasibility is internalized into the representation.

In this study, we investigate these mechanisms using a small traveling salesman problem (TSP) as an interpretable and fully analyzable testbed. 
We first validate that the bAE can reconstruct feasible tours with high fidelity. 
We then quantify how the learned binary latent space preserves the structure of the original solution space from both global and local perspectives, and compare it with handcrafted encodings under comparable compression rates. 
Finally, we evaluate the impact of the learned representation on FMQA performance in terms of approximation quality and the probability of generating feasible solutions. 
Through these analyses, we provide practical design insights into latent-representation-based black-box optimization and clarify how solution-space compression and structure-preserving latent geometry contribute to smoother and more feasible search landscapes.
Importantly, our goal is not to propose yet another optimization pipeline, but to make the learned representation itself interpretable and actionable for surrogate-based search. 

In a controlled setting, we evaluate reconstruction accuracy, structure preservation, and downstream optimization behavior side by side, and identify which geometric properties of the binary latent space are most strongly linked to feasibility and search efficiency.
We expect these findings to serve as practical guidance for designing latent representations and encodings in QUBO-based black-box optimization beyond the specific TSP instance studied here.

\section*{Results}
\label{sec:results}
In this section, we show that bAE can learn a binary latent representation that reflects the structure of the feasible-solution set of the TSP, and that QUBO optimization performed in this latent space is advantageous both in feasibility and in search efficiency. 
We present the results in the following order: reconstruction fidelity of the bAE, structure preservation in the learned latent space, and optimization performance within the FMQA cycle. 
Detailed definitions of the metrics and the full experimental setup are given in Methods. 
Here we summarize only what is needed to follow the Results.

\subsection*{bAE learns faithful binary reconstructions of feasible tours}
As a prerequisite for the downstream FMQA optimization, we first verify that the bAE can map feasible tours to a binary latent code and reconstruct them with high fidelity through the decoder. 
In our framework, the Ising machine searches for low-energy solutions of the QUBO defined in the latent space, and candidate tours are obtained by decoding the optimized latent codes. 
Therefore, the reconstruction performance of the bAE determines both the set of solutions that the search can practically reach and the stability of subsequent candidate evaluation.

In practical applications, the amount of available training data can vary substantially depending on problem size, constraint structure, and how feasible samples are generated. 
Establishing a universal sufficient dataset size is beyond the scope of this study. Instead, we prioritize analyzing the latent-space structure and downstream optimization behavior under conditions where underfitting due to data scarcity is minimized. 

Since the 8-city TSP has a finite solution space of $(L-1)!=5040$ feasible tours, we train the bAE on $5000$ tours, which covers almost the entire space.

We measure reconstruction loss using the mean squared error (MSE) between the input tour representation $\bm{\pi}$ and its reconstruction $\hat{\bm{\pi}}$ (see Methods for the exact definition). 
We also define reconstruction accuracy $\mathcal{A}$ as the fraction of samples whose reconstructed tour exactly matches the original tour. Since $\mathcal{A}$ counts a tour as incorrect even if a single position differs, it is a stringent metric that does not tolerate local errors.
Thus, an increase in $\mathcal{A}$ provides direct evidence that the latent code retains the global structure of the tour.

\begin{figure}[t]
  \begin{minipage}[b]{0.45\linewidth}
    \centering
    \includegraphics[keepaspectratio, scale=0.6]{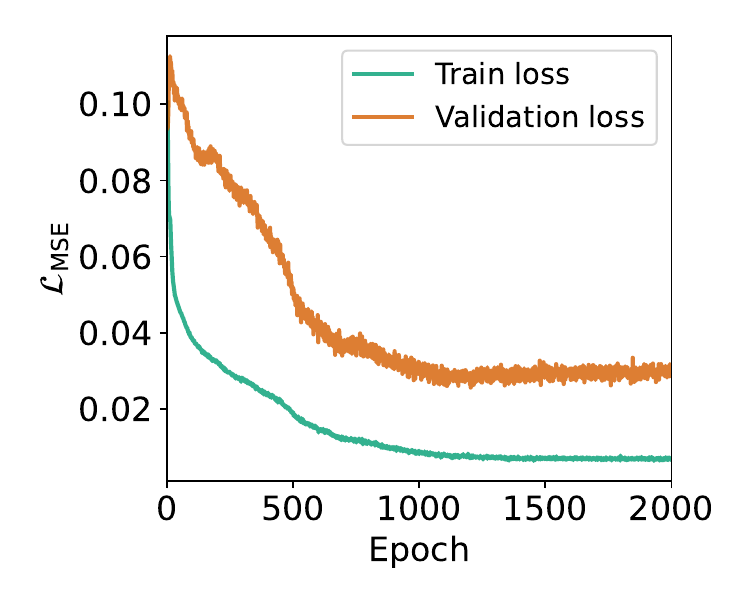}
  \end{minipage}
  \begin{minipage}[b]{0.45\linewidth}
    \centering
    \includegraphics[keepaspectratio, scale=0.6]{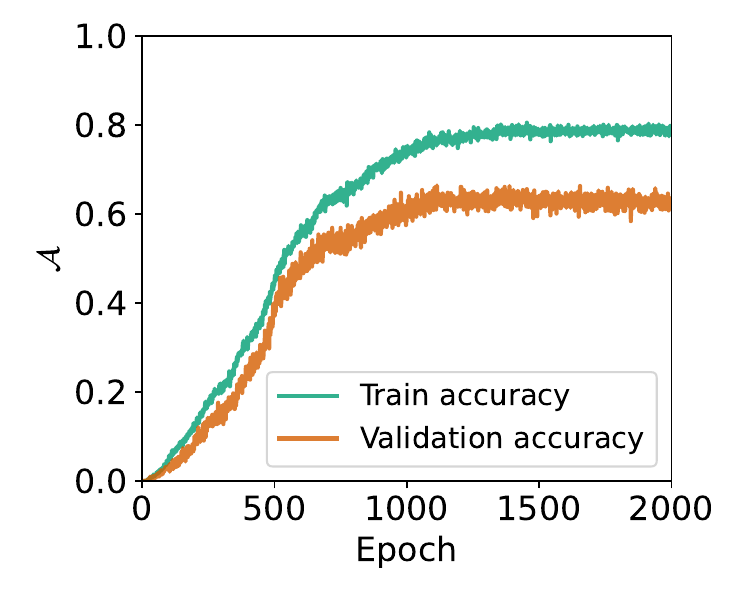}
  \end{minipage}
  \caption{Training curves of the bAE with $(d_z,d_h)=(14,64)$ on feasible tours. (a) Reconstruction loss $\mathcal{L}_{\mathrm{MSE}}$ as a function of epoch. (b) Reconstruction accuracy $\mathcal{A}$ as a function of epoch. Both training and validation metrics are shown.}
  \label{fig:bAE_loss_accuracy}
\end{figure}

Figure~\ref{fig:bAE_loss_accuracy} shows the training dynamics of the bAE in terms of the reconstruction loss $\mathcal{L}_{\mathrm{MSE}}$ and the reconstruction accuracy $\mathcal{A}$. 
As shown in Fig.~\ref{fig:bAE_loss_accuracy}(a), the reconstruction loss rapidly decreases in the early stage of training and then gradually converges. 
A similar decreasing trend is observed for the validation loss, and no clear increase is observed in the late stage of training, indicating the absence of severe overfitting under the present setting.
Consistent with the loss reduction, Fig.~\ref{fig:bAE_loss_accuracy}(b) shows that the reconstruction accuracy $\mathcal{A}$ increases during training and stabilizes at a high value in the later epochs.

Although the final training accuracy is slightly higher than the validation accuracy, suggesting mild overfitting, the bAE still generalizes to unseen tours and encodes global tour information in the binary latent space.

\begin{figure}[t]
  \begin{minipage}[b]{0.45\linewidth}
    \centering
    \includegraphics[keepaspectratio, scale=0.6]{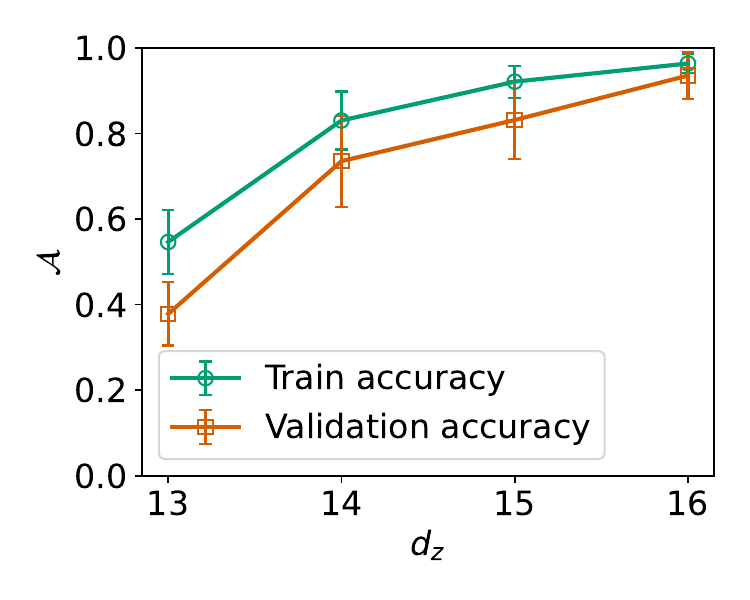}
  \end{minipage}
  \begin{minipage}[b]{0.45\linewidth}
    \centering
    \includegraphics[keepaspectratio, scale=0.6]{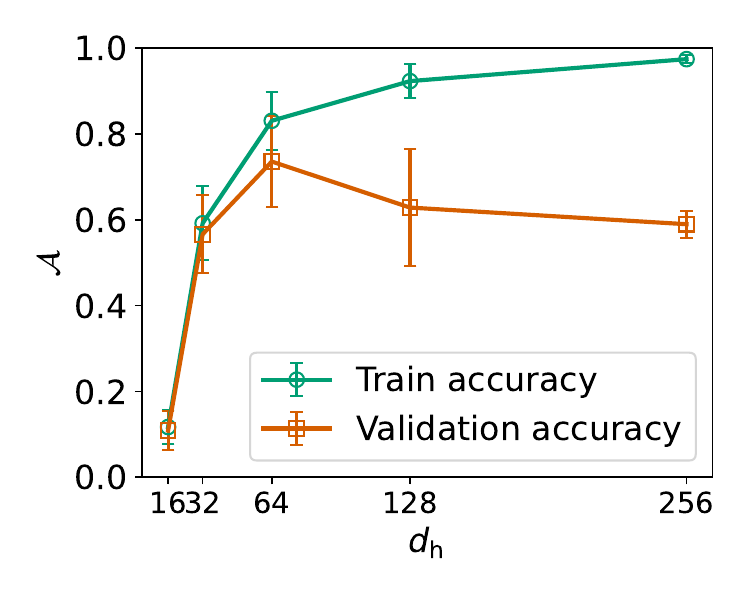}
  \end{minipage}
  \caption{Dependence of reconstruction accuracy on the latent dimension $d_z$ and the hidden-layer size $d_h$. (a) Reconstruction accuracy $\mathcal{A}$ versus $d_z$ with $d_h=64$. (b) Reconstruction accuracy $\mathcal{A}$ versus $d_h$ with $d_z=14$. 
  Each point shows the mean over five independent training runs with different random seeds. Error bars indicate the standard deviation. Lines connecting the points are drawn as a guide to the eye.}
  \label{fig:accuracy_vs_dz}
\end{figure}

Next, we determine a representative model configuration for subsequent analyses by examining the dependence of reconstruction accuracy on the latent dimension $d_z$ and the hidden-layer size $d_h$, which respectively control the compression rate and model capacity. Figure~\ref{fig:accuracy_vs_dz}(a) shows the dependence of $\mathcal{A}$ on $d_z$ with the hidden-layer size fixed. 
Each data point represents the average over five independent training runs with different random seeds, and the error bars indicate the standard deviation.
For small $d_z$, the reconstruction accuracy is low and exhibits large variance, whereas $\mathcal{A}$ improves and stabilizes as $d_z$ increases. 
However, increasing $d_z$ also exponentially enlarges the latent search space size $2^{d_z}$, which can make optimization on an Ising machine more difficult. 
Therefore, it is crucial to identify a regime where reconstruction fidelity saturates without introducing excessive dimensionality.
Based on this trade-off, we adopt $d_z = 14$ as the smallest representative value that yields sufficiently high and stable reconstruction accuracy, with approximately $70\%$ of tours reconstructed correctly on average. 
Figure~\ref{fig:accuracy_vs_dz}(b) further shows the dependence of $\mathcal{A}$ on the hidden-layer size $d_h$ with $d_z$ fixed at 14. 
While the training accuracy increases monotonically with $d_h$, the validation accuracy reaches its maximum around $d_h = 64$ and decreases for larger values, indicating overfitting. 
Accordingly, we use $(d_z, d_h) = (14, 64)$ as the representative configuration in the following analyses of latent-space structure and FMQA optimization.

\subsection*{Structure preservation in the latent space}
Here, we evaluate to what extent the binary latent space learned by the bAE (with the standard setting $(d_z, d_h) = (14, 64)$) preserves the structure of the original TSP solution space.
As baselines, we consider rank-based log/gray encoding, in which tours are first mapped to a rank-based representation and then binary encoded, as well as random label encoding, in which bit assignments are given randomly (Precise definitions of these encodings are provided in Methods).
If proximity in the latent Hamming space is consistent with proximity in the original tour space, the resulting search landscape becomes smoother, which is expected to facilitate exploration by an Ising machine. 
To quantitatively assess this structure-preservation property, we compare the encodings using three complementary metrics: the Spearman rank correlation of distances $\rho_{\mathrm{Spearman}}$, the neighborhood distance characteristic $L(m)$ under small bit flips, and the local optimum ratio $r_{\mathrm{Local}}$.

\begin{figure}[t]
  \begin{minipage}[b]{0.33\linewidth}
    \centering
    \includegraphics[keepaspectratio, scale=0.45]{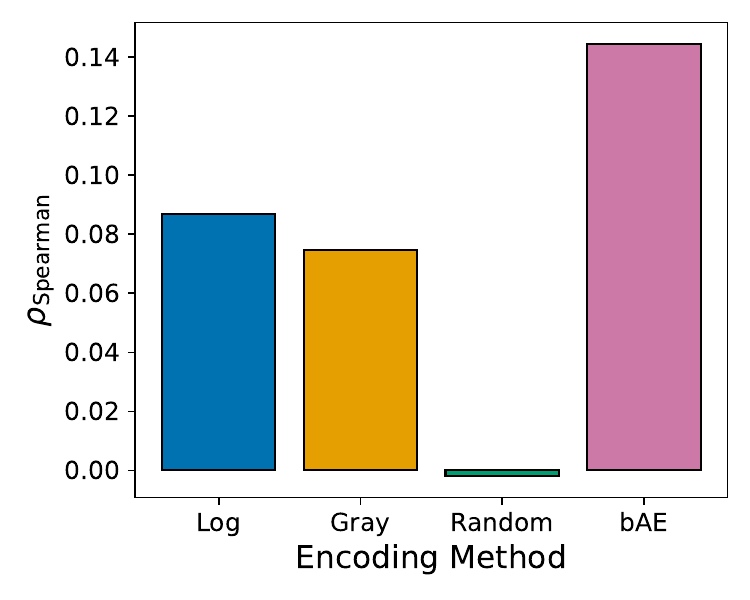}
  \end{minipage}
  \begin{minipage}[b]{0.33\linewidth}
    \centering
    \includegraphics[keepaspectratio, scale=0.45]{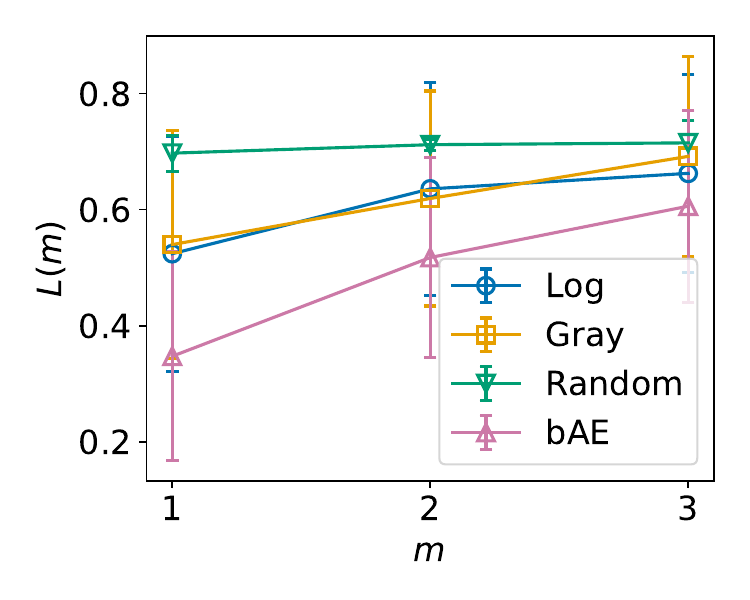}
  \end{minipage}
  \begin{minipage}[b]{0.33\linewidth}
    \centering
    \includegraphics[keepaspectratio, scale=0.45]{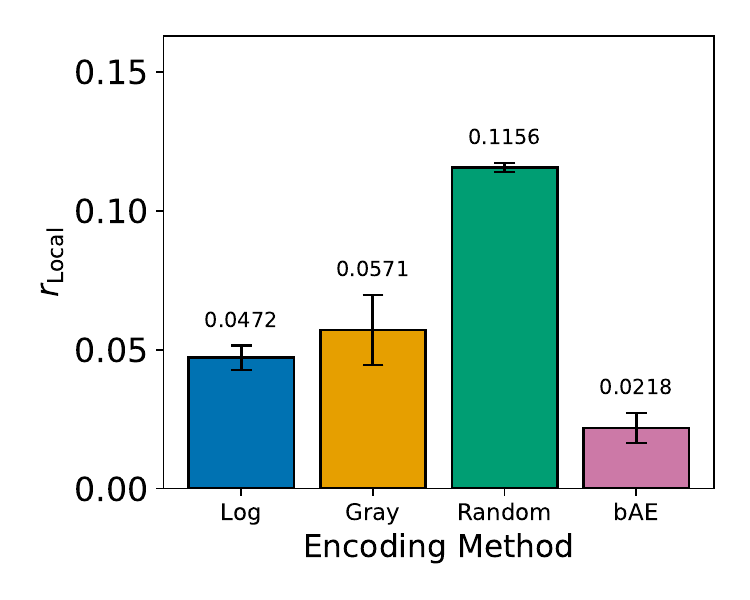}
  \end{minipage}
  \caption{Comparison of structure preservation in the binary space for the bAE, rank-based log/gray encoding, and random label encoding. (a) Spearman rank correlation coefficient between the edge distance in the original tour space and the Hamming distance in the binary space. (b) Neighborhood distance characteristic $L(m)$, which measures the change in tour distance induced by flipping $m$ bits in the binary space. (c) Local optimum ratio $r_{\mathrm{Local}}$, i.e., the fraction of binary solutions that are local optima under single-bit flips.}
  \label{fig:structure_preservation}
\end{figure}

We first examine how well the global distance structure of the solution space is preserved. 
Specifically, we compute the Spearman rank correlation coefficient $\rho_{\mathrm{Spearman}}$ between the edge distance $d_{\mathrm{Edge}}$ in the original tour space and the Hamming distance $d_{\mathrm{Hamming}}$ in the binary latent space (see Methods for definitions).
As shown in Fig.~\ref{fig:structure_preservation}(a), the bAE exhibits the highest positive correlation among all methods, indicating that it best preserves the relative ordering of "near" and "far" tours in the latent space. 
Rank-based log/gray encoding achieves a higher correlation than random label encoding but remains inferior to the bAE, while random label encoding yields a correlation close to zero, implying that the global structure is almost entirely destroyed. 
For random label encoding, the reported values are averaged over multiple encodings generated with different random seeds.

Next, we evaluate local neighborhood structure, which directly affects search efficiency. 
Since Ising machines perform energy-landscape-based exploration, it is desirable that small moves in the binary space correspond to small modifications in the original solution space. 
To quantify this property, we use the neighborhood distance characteristic $L(m)$, which measures how much the tour distance changes when transitioning to a point obtained by flipping $m$ bits in the binary space (see Methods).
As shown in Fig.~\ref{fig:structure_preservation}(b), random label encoding yields consistently large $L(m)$ with little dependence on $m$ indicating a weak relationship between binary distance and tour distance. Rank-based log/gray encodings show a similar trend. 
In contrast, for the bAE, $L(m)$ increases as $m$ increases. 
This $m$ dependence is largely absent in random label encoding and is more pronounced than in rank-based log/gray encodings over the range of $m$ considered.

Finally, we evaluate local structure from the perspective of the prevalence of local optima. 
We define a binary solution $\mathbf{z}$ as a local optimum if flipping any single bit does not improve the objective value, and we define the local optimum ratio $r_{\mathrm{Local}}$ as the fraction of such solutions in the entire space (see Methods). 
A higher $r_{\mathrm{Local}}$ implies a landscape with more dead ends, which is disadvantageous for annealing-based search.
As shown in Fig.~\ref{fig:structure_preservation}(c), random label encoding exhibits the highest local optimum ratio (mean $0.1156$), indicating that approximately 10\% of solutions are local optima. 
Rank-based log and gray encodings yield smaller values ($0.0472$ and $0.0571$, respectively) but still exceed that of the bAE.
The bAE achieves the lowest local optimum ratio (mean $0.0218$), indicating that the latent landscape contains relatively few local optima. 
This result is consistent with the improved neighborhood continuity observed in Fig.~\ref{fig:structure_preservation}(b).

Overall, these results demonstrate that the bAE learns a latent binary representation that is more suitable for search than handcrafted encodings, not only in terms of global distance-order consistency (Spearman correlation) but also in terms of local continuity under small bit flips and a reduced prevalence of local optima.

\subsection*{Solution-space compression enhances feasibility and search efficiency in FMQA}
\label{subsec:results_optimization}

We next examine how the structure-preserving latent space learned by the bAE translates into practical optimization performance within the FMQA cycle. 
In FMQA, a factorization machine (FM) is trained on previously evaluated solutions and their objective values, and the trained FM is cast as a QUBO and optimized by an Ising machine to propose new candidate solutions. 
Therefore, if a latent representation induces a search-friendly landscape, performance improvements should appear as (i) faster progress toward high-quality candidates (higher search efficiency) and (ii) more stable generation of feasible solutions in constrained problems (higher feasibility).

\begin{figure}[t]
  \begin{minipage}[b]{0.45\linewidth}
    \centering
    \includegraphics[keepaspectratio, scale=0.6]{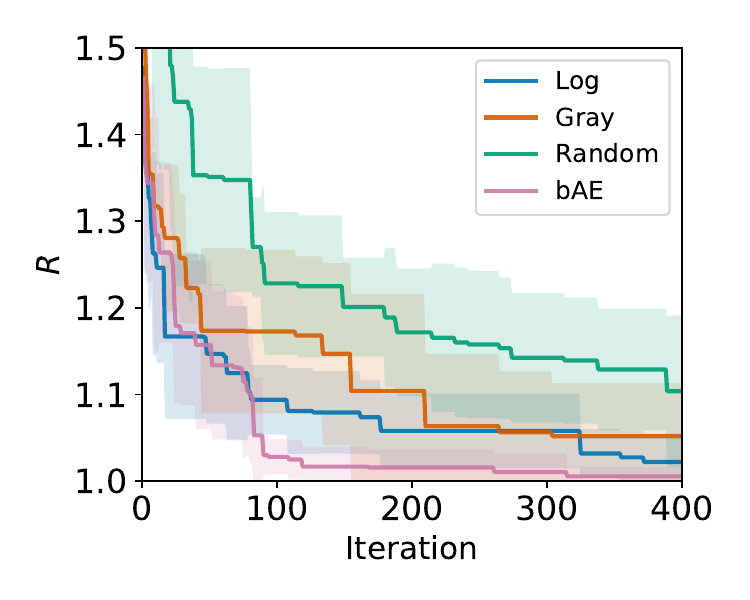}
  \end{minipage}
  \begin{minipage}[b]{0.45\linewidth}
    \centering
    \includegraphics[keepaspectratio, scale=0.6]{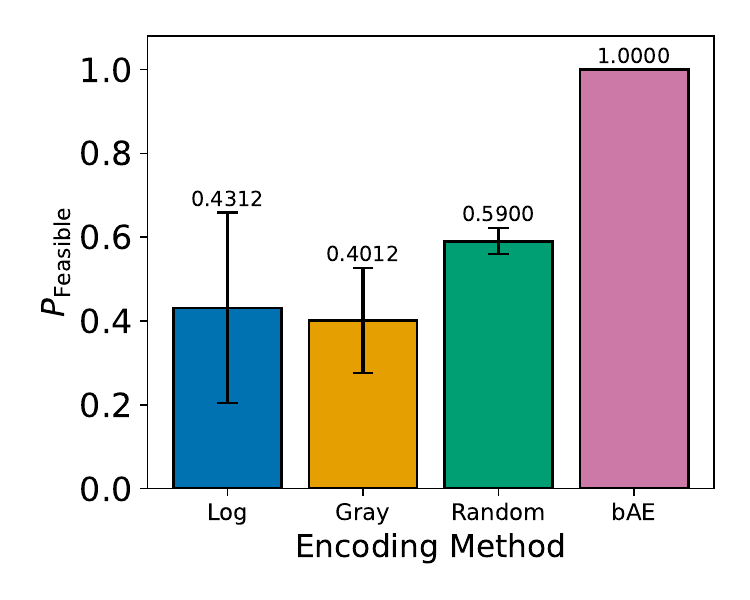}
  \end{minipage}
  \caption{Optimization performance of bAE+FMQA. (a) Approximation ratio $R$ as a function of FMQA iterations, where $R\approx1$ indicates reaching the exact optimum. (b) Feasible-sample probability $P_{\mathrm{Feasible}}$, defined as the fraction of raw Ising machine outputs that satisfy the constraints before any post-processing. Error bars indicate mean $\pm$ standard deviation over problem instances.}
  \label{fig:opt_bAE}
\end{figure}

To quantify optimization performance, we use the approximation ratio $R$ with respect to the exact optimum (see Methods), where smaller values indicate better solutions and faster decreases in $R$ across iterations indicate more efficient search. Figure~\ref{fig:opt_bAE}(a) shows the evolution of $R$ over FMQA iterations. Random label encoding exhibits only modest improvement, with $R$ remaining high throughout the search. 
Rank-based log/gray encoding performs better than random label encoding, yet it is consistently inferior to the bAE both in convergence speed and in the final attainable ratio.
In contrast, the bAE rapidly reduces $R$ from early iterations and reaches $R\approx 1$ with substantially fewer iterations. 
This indicates that iterative optimization combining FM-based quadratic approximation and Ising machine search operates most effectively on the bAE-induced latent landscape.

We further evaluate feasibility in constrained optimization using the feasible-sample probability $P_{\mathrm{Feasible}}$, defined as the fraction of raw Ising machine outputs that satisfy the constraints before any post-processing (see Methods). This metric directly reflects how strongly feasibility is internalized by the representation. Figure~\ref{fig:opt_bAE}(b) shows $P_{\mathrm{Feasible}}$ for each method. random label encoding yields $P_{\mathrm{Feasible}}\approx0.6$. Rank-based Log and Gray Encodings produce even lower $P_{\mathrm{Feasible}}$ values and exhibit noticeable instance-to-instance variability, indicating unstable feasibility. In sharp contrast, the bAE achieves $P_{\mathrm{Feasible}}=1.0$ in all trials, meaning that the search produces only feasible tours throughout the optimization process.

Taken together, these results demonstrate that solution-space compression via the bAE improves FMQA in two complementary ways: it accelerates search (a faster decrease in $R$) and robustly guarantees feasibility in constrained problems ($P_{\mathrm{Feasible}}=1.0$), thereby stabilizing and enhancing the overall FMQA cycle.

\section*{Discussion}
In this study, we have quantitatively elucidated why the bAE+FMQA framework functions effectively in combinatorial optimization, focusing on two key aspects: structure preservation in the latent space and solution-space compression. Rather than merely demonstrating performance gains, we explicitly connected geometric properties of the learned binary latent representation to downstream search behavior.
Specifically, we showed that the bAE can faithfully encode feasible TSP tours into a compact binary latent space while maintaining high reconstruction fidelity (Fig.~\ref{fig:bAE_loss_accuracy}). More importantly, the learned latent space exhibits superior structure preservation compared with handcrafted encodings, both in terms of global distance-order consistency, as measured by the Spearman rank correlation, and local neighborhood continuity under small bit flips, as captured by the neighborhood distance characteristic $L(m)$ (Fig.~\ref{fig:structure_preservation}(a,b)). The reduced local optimum ratio observed for the bAE (Fig.~\ref{fig:structure_preservation}(c)) further indicates that the induced energy landscape in the latent space is relatively smooth, with fewer dead ends that can trap annealing-based search.
These geometric properties directly translate into improved optimization behavior within the FMQA cycle. When combined with FM-based quadratic approximation and Ising machine search, the bAE-induced latent representation leads to a faster reduction in the approximation ratio $R$ and achieves the exact optimum with fewer iterations (Fig.~\ref{fig:opt_bAE}(a)).
At the same time, feasibility is dramatically improved: the bAE consistently achieves a feasible-sample probability $P_{\mathrm{Feasible}}=1.0$, meaning that all raw Ising machine outputs correspond to valid tours without requiring post-processing (Fig.~\ref{fig:opt_bAE}(b)).
This stands in sharp contrast to random and rank-based log/gray encodings, which exhibit both lower feasibility and substantial instance-to-instance variability.

These results suggest that at least two mechanisms are crucial for the success of bAE+FMQA. The first is structure preservation in the latent space. In handcrafted encodings such as random label encoding, even a single-bit flip can induce a large and essentially random change in the original solution, destroying locality and resulting in a rugged energy landscape with many local optima. Since Ising machines explore solutions by following the energy landscape defined by a QUBO, such mismatch between binary neighborhoods and meaningful solution similarity is inherently disadvantageous. In contrast, the bAE achieves the highest distance-rank correlation and the best neighborhood continuity, indicating that local moves in the latent space correspond more reliably to local modifications in the original tour space. 
This alignment enables the FM surrogate model to provide informative guidance and allows annealing-based search to proceed more smoothly.

The second mechanism is solution-space compression. In our setting, the bAE is trained exclusively on feasible solutions, which implicitly folds the feasible region of the original solution space into a compact latent manifold. As a result, candidate points sampled in the latent space are highly likely to decode into feasible solutions. The fact that $P_{\mathrm{Feasible}}=1.0$ is achieved throughout the optimization process suggests that feasibility is effectively internalized into the representation itself, rather than being enforced by explicit penalties or post-processing. By contrast, random label encoding yields a feasible-sample probability comparable to the fraction of feasible points in the original binary space, indicating that little feasibility information is captured by the encoding.
Importantly, this improvement cannot be attributed solely to dimensionality reduction. Simply reducing the number of bits does not guarantee either feasibility or efficient search, as evidenced by the inferior performance of handcrafted encodings with comparable compression rates. Instead, both compression and structure preservation must be simultaneously satisfied for FMQA to function effectively in constrained optimization problems.

From a broader perspective, our study is deliberately positioned at a small-scale and interpretable setting. The goal is not to claim immediate scalability to large TSP instances, but rather to isolate and analyze the mechanisms by which bAE+FMQA succeeds. By exhaustively characterizing the relationship between the original solution space and the learned latent geometry, we provide interpretable evidence linking reconstruction fidelity, distance preservation, neighborhood continuity, and downstream optimization performance.

These findings offer practical guidance for the design of latent representations in QUBO-based black-box optimization. Beyond reconstruction accuracy alone, desirable representations should preserve distance orderings, maintain locality under small bit flips, and reduce the prevalence of local optima in the induced energy landscape. Future work should investigate how such geometric properties can be encouraged through architectural choices, training objectives, or inductive biases tailored to specific problem structures and constraints. In addition, clarifying the relationship between training data size, diversity, and the resulting latent geometry remains an important open question.

Overall, by making the latent representation itself interpretable and actionable, this study advances understanding of how learned binary embeddings can systematically improve feasibility and search efficiency in surrogate-based optimization with Ising machines.

\section*{Methods}
This section describes the problem setting, data generation procedure, handcrafted baseline encodings, training of the binary autoencoder (bAE), evaluation metrics for latent-space structure, and the optimization protocol based on FMQA.

\subsection*{TSP instance and tour representation}
To analyze the behavior of the bAE+FMQA framework in an interpretable manner, we consider the traveling salesman problem (TSP) as a representative constrained combinatorial optimization task. 
We focus on a small-scale setting with the number of cities fixed to $L=8$.
City coordinates $\{\bm{r}_i\}_{i=1}^{L}$ are independently sampled from a uniform distribution over the unit square $[0,1]\times[0,1]$. 
The distance between cities $i$ and $j$ is defined as the Euclidean distance $d_{ij}=\|\bm{r}_i-\bm{r}_j\|_2$. A tour is represented as an ordered sequence of cities $\boldsymbol{\pi}=(\pi_1,\ldots,\pi_L)$, and the objective function to be minimized is given by
\begin{equation}
    f(\bm{\pi})=\sum_{t=1}^{L} d_{\pi_t,\pi_{t+1}},\quad \pi_{L+1}\equiv \pi_1.
\end{equation}
To remove rotational symmetry, the starting city $\pi_1$ is fixed in the representation, resulting in $(L-1)!$ feasible tours. Reverse-order tours are not identified and are treated as distinct solutions for simplicity of analysis.

The training dataset for the bAE is constructed by uniformly sampling $N_{\mathrm{total}}$ feasible tours from this solution set. The dataset is split into training and validation subsets with an $4{:}1$ ratio to evaluate reconstruction performance. Owing to the small problem size, the exact optimal tour length $f^\star$ can be computed by exhaustive enumeration and is used as a reference for optimization-performance metrics.
In practical applications, the amount of available training data depends strongly on problem size, constraint structure, and data-generation mechanisms. Establishing a generally sufficient dataset size is therefore beyond the scope of this study. Instead, we prioritize eliminating underfitting due to data scarcity and analyze latent-space structure and downstream optimization behavior under conditions where the model performance is sufficiently saturated. Specifically, we exploit the finiteness of the $L=8$ TSP solution space and train the bAE using a relatively large dataset ($N_{\mathrm{total}}=5000$).

\subsection*{Binary autoencoder (bAE)}
We employ a binary autoencoder (bAE) to map a feasible tour $\boldsymbol{\pi}$ to a low-dimensional binary latent variable $\boldsymbol{z}\in \{0,1\}^{d_z}$ and to reconstruct a tour $\hat{\boldsymbol{\pi}}$ from $\boldsymbol{z}$. 
An autoencoder (AE)~\cite{hinton2006reducing} is an unsupervised neural network that compresses input data into a latent space and is trained to reconstruct the input from the latent representation. AEs whose latent variables are explicitly binary are often referred to as binary AEs (bAEs)~\cite{baynazarov2019binary}. 
In our setting, the goal of the bAE is to learn a compact binary representation that is aligned with the manifold of feasible solutions.

Following prior work, we implement a sequence-to-sequence (Seq2Seq) model based on the Gated Recurrent Unit (GRU)~\cite{cho2014learning}. Since a TSP tour is naturally represented as a sequence (permutation) whose length depends on the number of cities $L$, a recurrent encoder--decoder architecture provides a convenient and interpretable baseline.
Let $\mathrm{Enc}(\cdot)$ and $\mathrm{Dec}(\cdot)$ denote the encoder and decoder, respectively:
\begin{align}
    \bm{z} &= \text{Enc}(\bm{\pi}) \in \{ 0,1 \}^{d_z}, \label{eq:encoder} \\ 
    \hat{\bm{\pi}} &= \text{Dec}(\bm{z}) \in \{ 1,2,\cdots, L \}^{L}. \label{eq:decoder}
\end{align}
Here $d_z$ is the latent dimension, and we aim for $d_z \ll L^2$ (for $L=8$, $L^2=64$), i.e., a substantially more compact representation than the original two-way one-hot tour matrix.
A tour is represented as an ordered city sequence $\boldsymbol{\pi}=(\pi_1,\ldots,\pi_L)$. At each time step $t$, we encode the city index $\pi_t$ as a one-hot vector $\boldsymbol{x}_t := \mathrm{onehot}(\pi_t)$.

The encoder processes the input sequence $\{\bm{x}_t\}_{t=1}^L$ with a GRU:
\begin{equation}
\bm{h}_t=\mathrm{GRU}(\bm{x}_t,\bm{h}_{t-1}).
\end{equation}
From the final hidden state $\boldsymbol{h}_{L}$, a fully connected layer with a sigmoid activation produces a continuous vector $\boldsymbol{p}\in[0,1]^{d_z}$:
\begin{equation}
    \bm{p} = \sigma(\bm{W} \bm{h}_{L} + \bm{b}),
\end{equation}
where $W$ and $\boldsymbol{b}$ are trainable parameters and $\sigma(\cdot)$ is the element-wise sigmoid function.
To obtain a binary latent vector, we apply stochastic thresholding:
\begin{equation}
    z_i=\mathrm{step}(p_i-\xi_i), \qquad \xi_i\sim U(0,1),
    \label{eq:binary_step}
\end{equation}
where $\mathrm{step}(\cdot)$ is the Heaviside step function and $U(0,1)$ denotes the uniform distribution. 
We denote the above pipeline (GRU $\rightarrow$ sigmoid $\rightarrow$ stochastic binarization) as $\mathrm{Enc}(\cdot)$ in Eq.~\eqref{eq:encoder}.

The decoder reconstructs a tour sequence from $\boldsymbol{z}$ using another GRU. First, the latent vector is mapped to the initial hidden state:
\begin{equation}
    \bm{h}_0 = \bm{W}_{\text{init}}\bm{z} + \bm{b}_{\text{init}}.
\end{equation}
At each time step $t$, the hidden state is updated by
\begin{equation}
    \bm{h}_t=\mathrm{GRU}(\tilde{\bm{x}}_{t-1},\,\bm{h}_{t-1}),\quad
    \tilde{\bm{x}}_{t-1}=
    \begin{cases}
    \bm{x}_{t-1} & (\text{training with teacher forcing})\\
    \hat{\bm{x}}_{t-1} & (\text{inference})
    \end{cases}
\end{equation}
where $\hat{\bm{x}}_{t-1}$ is the one-hot vector corresponding to the previously predicted city. 
A linear layer produces a score vector $\boldsymbol{u}_t\in\mathbb{R}^{L}$:
\begin{equation}
    \bm{u}_t = \bm{W}_0 \bm{h}_t + \bm{b}_0,
\end{equation}
and the predicted distribution is obtained by
\begin{equation}
    \bm{p}_t=\mathrm{softmax}(\bm{u}_t).
\end{equation}
During inference, we choose the most likely city at each step:
\begin{equation}
    \hat{\pi}_t=\arg\max_{i\in\{1,\ldots,L\}}\, p_{t,i},
\end{equation}
and define $\hat{\boldsymbol{x}}_t=\mathrm{onehot}(\hat{\pi}_t)$. We denote the entire mapping $\boldsymbol{z}\mapsto \hat{\boldsymbol{\pi}}$ as $\mathrm{Dec}(\cdot)$ in Eq.~\eqref{eq:decoder}.

To train the bAE, we uniformly sample $N_{\mathrm{total}}$ feasible tours from the TSP solution space and split them into training and validation sets with $N_{\mathrm{train}}$ and $N_{\mathrm{valid}}$ samples, respectively.
\begin{equation}
    \mathcal{L}_\text{MSE} = \frac{1}{N_{\mathrm{train}}}\sum_{n=1}^{N_{\mathrm{train}}}\frac{1}{L}
    \sum_{t=1}^{L}\sum_{i=1}^{L}\left(p^{(n)}_{t,i}-x^{(n)}_{t,i}\right)^2.
    \label{eq:mse_loss}
\end{equation}
Since the binarization in Eq.~\eqref{eq:binary_step} is non-differentiable, we adopt the straight-through estimator~\cite{bengio2013estimating}.
In backpropagation, we approximate the gradient through the step function by
\begin{equation}
    \frac{\partial \mathcal{L}_{\mathrm{MSE}}}{\partial p_i}
    \coloneqq
    \frac{\partial \mathcal{L}_{\mathrm{MSE}}}{\partial z_i},
\end{equation}
where $\mathcal{L}_{\mathrm{MSE}}$ is defined in Eq.~\eqref{eq:mse_loss}.

\subsection*{Baseline encodings}

To assess the effectiveness of the latent representations learned by the bAE, we compare them with several handcrafted binary encodings. 
The purpose of this comparison is to evaluate how well a binary representation preserves the neighborhood structure of the original solution space.

Accordingly, we restrict our comparison to encodings that map a TSP tour directly to a binary vector. 
We consider three representative baselines: rank-based Log, rank-based gray, and random label. 
Rank based Log achieves high compression but does not explicitly preserve geometric neighborhood relations. 
Rank based Gray enforces algebraic locality by ensuring that adjacent ranks differ by a single bit. 
Random Label is expected to destroy the original structure.

All three encodings reduce the original TSP representation, which would require $L^2=64$ binary variables in a two-way one-hot formulation, to the minimum number of bits $B$ required to represent all feasible tours. 
In our setting, the number of feasible tours is $(L-1)!$, and thus the required bit length is given by $B=\lceil \log_2 (L-1)! \rceil = 13$. 
Consequently, all baseline encodings represent tours as binary vectors of length $B=13$.
Conventional encodings such as one-hot representations are excluded from this comparison, as they correspond to the original uncompressed representation rather than a compressed binary encoding. Below, we describe each baseline encoding in detail.

Rank-based Log encoding represents a permutation using the information-theoretically minimal number of bits based on its lexicographic rank. Specifically, a tour $\boldsymbol{\pi}$ with $L$ cities is mapped to an integer $R(\boldsymbol{\pi}) \in \{0,1,\ldots,(L-1)!-1\}$ using the Lehmer code (factorial number system)~\cite{Lehmer1960TeachingCT}. 
This integer is then converted into a $B$-bit binary vector:
\begin{equation}
\boldsymbol{z} = \mathrm{bin}_B \bigl(R(\boldsymbol{\pi})\bigr),
\end{equation}
where $\mathrm{bin}_B(\cdot)$ denotes the $B$-bit binary representation. While this encoding achieves maximal compression efficiency, numerical proximity in lexicographic rank does not necessarily correspond to geometric similarity between tours, such as shared edges.

Rank-based Gray encoding applies a Gray code transformation to the lexicographic rank $R(\boldsymbol{\pi})$. In contrast to standard binary representations, Gray codes ensure that adjacent integers differ by exactly one bit. The encoding is defined as
\begin{equation}
\boldsymbol{z} = \mathrm{gray}_B \bigl(R(\boldsymbol{\pi})\bigr), \qquad
\mathrm{gray}(R) = R \oplus (R \gg 1),
\end{equation}
where $\oplus$ denotes bitwise XOR and $\gg$ denotes the right shift. If lexicographically adjacent permutations were also geometrically similar, this encoding would preserve local neighborhood structure.
However, such correspondence is not guaranteed for permutations.

As a structure-destroying baseline, we also consider Random Label encoding. In this scheme, each feasible tour $\boldsymbol{\pi}$ is assigned a unique random integer label, which is then converted into a $B$-bit binary vector. This procedure is equivalent to randomly permuting the coordinates of the solution space and is therefore expected to completely destroy any meaningful neighborhood or distance structure.

For all handcrafted encodings, decoding is performed by mapping a binary vector $\boldsymbol{z}$ back to an integer rank $R(\boldsymbol{\pi})$ and reconstructing the corresponding tour. Since the representable integer range $2^B$ generally exceeds $(L-1)!$, decoded values satisfying $R(\boldsymbol{\pi}) \ge (L-1)!$ are treated as invalid codes and are regarded as constraint violations during optimization and evaluation.

\subsection*{QUBO approximation on latent space using factorization machine}
Using the bAE or the baseline encodings described above, each feasible TSP tour $\boldsymbol{\pi}$ is mapped to a binary vector $\boldsymbol{z} \in \{0,1\}^{d_{\mathrm{bin}}}$, where $d_{\mathrm{bin}}=d_z$ for the bAE and $d_{\mathrm{bin}}=B$ for the handcrafted encodings.
If the tour length $f(\boldsymbol{\pi})$ can be approximated as a function defined on this latent space, the optimization problem can be reformulated as a QUBO whose variables are only the latent binary variables $\boldsymbol{z}$. 
This enables direct optimization using an Ising machine.

In this study, we adopt factorization machine (FM)~\cite{rendle2010factorization} as the surrogate model for approximating the objective function on the latent space. 
An FM is a regression model that represents first-order terms and all pairwise interactions of the input variables through inner products of low-rank factor vectors. 
This formulation naturally induces a quadratic form for binary inputs.
Specifically, for a latent binary vector $\boldsymbol{z} \in \{0,1\}^{d_{\mathrm{bin}}}$, the FM approximates the tour length as
\begin{equation}
  f_{\text{FM}}(\bm{z}) = w_0 + \sum_{i=1}^{d_{\mathrm{bin}}} w_i z_i 
  + \sum_{i<j} \langle \bm{v}_i, \bm{v}_j \rangle z_i z_j.
  \label{eq:12_fm_model}
\end{equation}
Since the latent variables $z_i$ are binary, Eq.~\eqref{eq:12_fm_model} is isomorphic to a QUBO energy function of the form $E(\boldsymbol{z}) = \sum_i Q_{ii} z_i + \sum_{i<j} Q_{ij} z_i z_j + \mathrm{const}$.
Indeed, by identifying $Q_{ii}=w_i$ and $Q_{ij}=\langle \boldsymbol{v}_i,\boldsymbol{v}_j\rangle$ for $i<j$, the trained FM parameters directly define the linear and quadratic coefficients of the corresponding QUBO, with the constant term given by $w_0$.

\subsection*{Factorization Machine with Quantum Annealing}
We now describe the hybrid optimization algorithm that integrates solution-space compression by the bAE (or baseline encodings) with objective-function approximation using FM, and performs optimization via an Ising machine. 
This framework is referred to as FMQA.

Given a trained bAE or a handcrafted encoding, FMQA proceeds through an iterative optimization cycle.
At the beginning of the process, a small set of feasible solutions and their corresponding objective values is prepared as an initial dataset $\mathcal{D}_{\text{init}}$.
Each solution is encoded into a binary vector $\boldsymbol{z}$ using the selected encoding scheme.
Using the encoded dataset ${(\boldsymbol{z}^{(m)}, f(\boldsymbol{\pi}^{(m)}))}$, an FM surrogate model is trained to approximate the objective function on the latent space, as described in the previous subsection. 
The learned FM parameters define a QUBO, which is then provided to an Ising machine to generate candidate solutions by sampling low-energy configurations.
The binary samples output by the Ising machine are decoded back into candidate tours in the original solution space using the corresponding decoding rule of the encoding scheme. The decoded solutions are evaluated using the true objective function, and the resulting pairs are added to the dataset. This expanded dataset is then used to retrain the FM, and the procedure is repeated.
Formally, the FMQA optimization cycle consists of the following steps:
\begin{enumerate}
    \item Prepare an initial dataset $\mathcal{D}_{\mathrm{init}}$ consisting of a small number of solutions and their objective values.
    \item Encode each solution into a binary vector using the trained bAE or a baseline encoding.
    \item Train an FM surrogate model on the encoded dataset.
    \item Construct a QUBO from the trained FM and sample candidate solutions using an Ising machine.
    \item Decode the sampled binary vectors into candidate solutions in the original space.
    \item Evaluate the decoded solutions and add them to the dataset.
    \item Repeat steps 2 - 6 until a termination criterion is satisfied.
\end{enumerate}

\subsection*{Settings}

The hyperparameter settings for the bAE and the FM are summarized in Table~\ref{tab:settings_bAE_FM}. 
The architectural choices and training parameters were determined through the hyperparameter search described in the Results, and the selected configuration was used for all subsequent analyses and optimization experiments. 
For parameter updates, we used AdamW~\cite{loshchilov2017decoupled}.
To optimize the QUBO constructed by the FM on the latent space, we used a quantum annealing machine, \textit{D-Wave Advantage system 6.4}. 
The annealing time was fixed to $20~\mu\mathrm{s}$ (default setting), and when minor embedding onto the hardware graph was required, we used the automatic embedding provided by the system.
\begin{table}[t]
    \centering
    \caption{Hyperparameter settings for the binary autoencoder (bAE) and the factorization machine (FM).}
    \begin{tabular}{|l|l|l|l|}
    \hline
    Model & Parameter & Value & Search range \\
    \hline
    bAE & Architecture & GRU & Fixed \\
    & Hidden dimension $d_h$ & 64 & 16 - 256 \\
    & Latent dimension $d_z$ & 14 & 13 - 16 \\
    & Number of encoder/decoder layers & 2 & Fixed \\
    & Learning rate & 0.001 & Fixed \\
    & Number of epochs & 2000 & Fixed \\
    \hline
    FM & Rank $k$ & 8 & Fixed \\
    & Learning rate & 0.01 & Fixed \\
    & Number of epochs & 1000 & Fixed \\
    & Initial dataset size $N_\text{init}$ & 100 & Fixed \\
    \hline
    \end{tabular}
   \label{tab:settings_bAE_FM}
\end{table}

\subsection*{Sample acceptance and post-processing in FMQA}
In FMQA, samples generated by the Ising machine are sequentially decoded, and the resulting solutions are added to the training dataset to advance the optimization process. However, indiscriminately accepting infeasible solutions or solutions that duplicate existing data can degrade the quality and diversity of the training set, leading to poorer generalization of the FM surrogate model and inefficient exploration.
To address this issue, we impose two conditions on each sample generated during the optimization cycle: (i) whether the decoded solution satisfies the problem constraints, and (ii) whether it is distinct from the existing dataset. If both conditions are satisfied, the solution is directly added to the dataset. Otherwise, post-processing procedures described below are applied.

\subsubsection*{Handling of new samples in FMQA}
This subsection describes how binary samples output by the Ising machine are post-processed when they do not correspond to valid tours after decoding. For each encoding scheme, we define a decoding post-processing rule that guarantees the final output is always a valid tour (i.e., a permutation of cities).
Specifically, we apply: (i) rank wrapping for rank-based encodings (Log and Gray), (ii) nearest-neighbor substitution for Random Label encoding, and (iii) deterministic repair for the bAE-based decoding.

For Log and Gray encodings, the binary string is decoded into an integer rank $r$, which is then mapped to a permutation (tour). Since the decoded rank may fall outside the valid range, we apply a wrapping operation
\begin{equation}
r_{\mathrm{wrap}} = r \bmod N_{\mathrm{codes}},
\end{equation}
and decode using $r_{\mathrm{wrap}}$. Here, $N_{\mathrm{codes}}$ denotes the total number of feasible tours, which for a TSP with $L$ cities (with rotational symmetry fixed) is given by
\begin{equation}
N_{\mathrm{codes}} = (L-1)!.
\end{equation}
This procedure ensures that a valid tour is always returned, even when the decoded rank is out of range.

In Random Label encoding, the binary string is decoded into an integer label $\ell$, which is then mapped back to a tour using a lookup table. When the decoded label is not present in the table, we compute the Hamming distance to all registered codes and substitute the tour corresponding to the nearest neighbor
\begin{equation}
    \ell^\ast = \underset{\ell' \in \mathcal{C}_{\mathrm{pool}}} {\arg\min}\quad d_{\mathrm{Hamming}}(\ell, \ell').
\end{equation}
Here, $\mathcal{C}_{\mathrm{pool}}$ denotes the fixed set of binary codes corresponding to all feasible tours used to define the Random Label encoding.
If multiple nearest neighbors exist, one is selected randomly. 
This operation can be interpreted as projecting an unseen label onto the closest known tour in the binary space.

For the bAE, the binary vector is passed through the decoder to obtain a city sequence. The decoded output may violate constraints due to duplicated or missing cities. In such cases, we apply a deterministic repair procedure consisting of two steps: (i) removing duplicated elements and out-of-range indices while preserving their order, and (ii) appending missing cities in ascending order. This guarantees construction of a valid permutation of length $L$.

\subsubsection*{Handling duplicate solutions in FMQA}
This subsection describes how FMQA handles decoded solutions that duplicate existing entries in the dataset. After decoding, each candidate tour is checked against the current dataset. If it is not a duplicate, it is directly added.
When duplication is detected, we perform a local search around the decoded tour to generate a new candidate. Specifically, we apply neighborhood operations consisting of (i) 2-opt moves, which remove two edges and reconnect the tour by reversing a subpath, and (ii) city-swap moves, which exchange the positions of two cities. These operations are attempted up to a maximum of 50 trials.
If a newly generated tour is not present in the dataset, it is accepted and added. If all attempts fail to produce a novel solution, the sample is discarded. This procedure prevents the dataset from collapsing onto a small set of repeated solutions and helps maintain diversity, thereby stabilizing FM training and improving exploration efficiency in FMQA.

\subsection*{Evaluation Metrics}
To comprehensively evaluate the proposed framework, we employ the following metrics, which quantify reconstruction fidelity, structural preservation of the solution space, landscape smoothness, and optimization performance.

\subsubsection*{Reconstruction loss}
The reconstruction loss measures how accurately the bAE reproduces the input tour representation. It is defined as the mean squared error (MSE) between the ground-truth one-hot sequence $\{\bm{x}_t\}_{t=1}^{L}$ derived from the input tour $\bm{\pi}$ and the probability distributions $\{\bm{p}_t\}_{t=1}^{L}$ output by the decoder:
\begin{equation}
    \mathcal{L}_{\mathrm{MSE}}
    =\frac{1}{N_{\mathrm{data}}}\sum_{n=1}^{N_{\mathrm{data}}}\frac{1}{L}
    \sum_{t=1}^{L}\sum_{i=1}^{L}\left(p^{(n)}_{t,i}-x^{(n)}_{t,i}\right)^2.
\end{equation}

\subsubsection*{Reconstruction accuracy}
Reconstruction accuracy is defined as the fraction of input tours that are perfectly reconstructed after decoding. 
A decoded tour is regarded as correct only if it exactly matches the input tour.
Even a single mismatch is counted as incorrect.
The reconstruction accuracy is given by
\begin{equation}
    \mathcal{A} = \frac{N_\text{accurate}}{N_\text{data}}.
\end{equation}
where $N_{\mathrm{accurate}}$ denotes the number of perfectly reconstructed tours.

\subsubsection*{Rank correlation coefficient}
To assess how well an encoding preserves the global structure of the solution space, we evaluate the rank correlation between distances in the original TSP space and distances in the corresponding binary representation. Specifically, for a pair of tours $\bm{\pi}^{(a)}$ and $\bm{\pi}^{(b)}$, we compute the edge distance $d_{\mathrm{Edge}}^{(a,b)}$ in the solution space and the normalized Hamming distance $d_{\mathrm{Hamming}}^{(a,b)}$ between their binary representations $\bm{z}^{(a)}$ and $\bm{z}^{(b)}$.
Let $d_{\mathrm{bin}}$ denote the bit length of the binary representation ($d_{\mathrm{bin}}=d_z$ for the bAE and $d_{\mathrm{bin}}=B$ for handcrafted encodings). The normalized Hamming distance is defined as
\begin{equation}
    d_{\mathrm{Hamming}}^{(a,b)} = \frac{1}{d_{\mathrm{bin}}} \sum_{q=1}^{d_{\mathrm{bin}}}\left| z_q^{(a)} -z_q^{(b)} \right|.
\end{equation}
The edge set of a tour $\bm{\pi}=(\pi_1,\ldots,\pi_L)$ is defined as
\begin{equation}
\mathcal{E}(\bm{\pi}) := \bigl\{ \{\pi_i,\pi_{i+1}\} \,\big|\, i=1,\ldots,L \bigr\}, \quad \pi_{L+1}=\pi_1,
\end{equation}
and the edge distance between two tours is given by
\begin{equation}
d_{\mathrm{Edge}}^{(a,b)} = 1-\frac{1}{L}\left|\mathcal{E}\!\left(\bm{\pi}^{(a)}\right)\cap \mathcal{E}\!\left(\bm{\pi}^{(b)}\right)\right|.
\end{equation}
Let $D_{\mathrm{H}}=\bigl(d_{\mathrm{Hamming}}^{(a,b)}\bigr)$ and
$D_{\mathrm{E}}=\bigl(d_{\mathrm{Edge}}^{(a,b)}\bigr)$ denote the lists of pairwise
Hamming distances and edge distances, respectively, computed over randomly sampled tour pairs.
We compute the Spearman rank correlation coefficient between $D_{\mathrm{H}}$ and $D_{\mathrm{E}}$ as
\begin{equation}
\rho_{\mathrm{Spearman}}
= \mathrm{corr}\!\left(\mathrm{rank}(D_{\mathrm{H}}),\,\mathrm{rank}(D_{\mathrm{E}})\right).
\end{equation}
Here, $\mathrm{rank}(\cdot)$ denotes ranking with averaged ties, and $\mathrm{corr}(\cdot,\cdot)$ denotes the Pearson correlation.

\subsubsection*{Neighborhood distance characteristic}
To quantify the smoothness of the energy landscape induced by an encoding, we analyze how small perturbations in the latent space affect the decoded solution. For an evaluation tour $\bm{\pi}$, its latent binary code $\bm{z}=\mathrm{Enc}(\bm{\pi})\in \{0,1\}^{d_{\mathrm{bin}}}$ is obtained. We then generate neighboring codes $\bm{z}'$ by flipping $m$ randomly selected bits such that $d_{\mathrm{Hamming}}(\bm{z},\bm{z}')=m$.
Each $\bm{z}'$ is decoded into a tour $\bm{\pi}'=\mathrm{Dec}(\bm{z}')$. When $\bm{\pi}'$ is feasible, we compute the edge distance $d_{\mathrm{Edge}}(\bm{\pi},\bm{\pi}')$. Repeating this process over multiple samples, we define
\begin{equation}
L(m)=\mathbb{E}_{\bm{\pi}}\Bigl[
\mathbb{E}_{\bm{z}'\,:\,d_{\mathrm{Hamming}}(\bm{z},\bm{z}')=m}
\bigl[
d_{\mathrm{Edge}}(\bm{\pi},\mathrm{Dec}(\bm{z}'))
\bigr]
\Bigr].
\end{equation}

\subsubsection*{Local optimum ratio}
To evaluate whether search in the latent space tends to be trapped by local optima, we measure the proportion of local optima among all possible solutions. A binary configuration $\bm{z}$ is defined as a local optimum if flipping any single bit does not improve the objective value:
\begin{equation}
\bm{z}\ \text{is a local optimum} \iff \forall \bm{z}'\in \mathcal{N}(\bm{z}),\ E(\bm{z}') \ge E(\bm{z}),
\end{equation}
where $\mathcal{N}(\bm{z})$ denotes the set of Hamming-1 neighbors of $\bm{z}$.
For the 8-city TSP considered in this study, all $(8-1)! = 5040$ feasible tours can be enumerated, allowing exact identification of local optima. The local optimum ratio is defined as
\begin{equation}
    r_{\mathrm{Local}}=\frac{N_{\mathrm{local}}}{N_{\mathrm{all}}},
\end{equation}
where $N_{\mathrm{local}}$ is the number of local optima and $N_{\mathrm{all}}=5040$ is the total number of feasible tours.

\subsubsection*{Approximation ratio}
To evaluate optimization performance, we use the approximation ratio $R$ with respect to the known optimal tour length $f^\ast$. 
Let $f$ denote the best objective value obtained during the optimization process. The approximation ratio is defined as
\begin{equation}
    R = \frac{f}{f^\ast}.
\end{equation}
Since the TSP is a minimization problem, $R \ge 1$, and values closer to $1$ indicate better solutions.

\subsubsection*{Feasible-sample probability}
In constrained combinatorial optimization, it is crucial that feasible solutions are generated consistently during the search process. We therefore define the feasible-sample probability $P_{\mathrm{Feasible}}$ as the fraction of raw samples output by the Ising machine that satisfy the TSP constraints (i.e., form valid tours) before any post-processing.
Let $N_{\mathrm{samp}}$ be the total number of samples generated by the Ising machine and $N_{\mathrm{Feasible}}$ the number of feasible ones. Then
\begin{equation}
    P_{\mathrm{Feasible}}=\frac{N_{\mathrm{Feasible}}}{N_{\mathrm{samp}}}.
\end{equation}
A higher value of $P_{\mathrm{Feasible}}$ indicates that the search remains closer to the feasible region and avoids inefficient exploration caused by constraint violations.

\bibliography{sample}

\section*{Acknowledgements}
This work was partially supported by the Japan Society for the Promotion of Science (JSPS) KAKENHI (Grant Number JP23H05447), the Council for Science, Technology, and Innovation (CSTI) through the Cross-ministerial Strategic Innovation Promotion Program (SIP), ``Promoting the application of advanced quantum technology platforms to social issues'' (Funding agency: QST), Japan Science and Technology Agency (JST) (Grant Number JPMJPF2221). In addition, this work is partially based on results obtained from a project, JPNP23003, commissioned by the New Energy and Industrial Technology Development Organization (NEDO).
S. Tanaka wishes to express their gratitude to the World Premier International Research Center Initiative (WPI), MEXT, Japan, for their support of the Human Biology-Microbiome-Quantum Research Center (Bio2Q).
T.~A. and M.~Y. thank Keita Takahashi for insightful discussions and valuable suggestions that helped improve this work.
Finally, they also thank the MITOU Target project for providing a development environment for annealing machines, and its project managers, Ryo Tamura and Kotaro Tanahashi, for meaningful discussions and comments. 
We further thank Yosuke Mukasa for technical support and helpful feedback as a technical advisor.

\section*{Author contributions statement}
T.~A. implemented the model, conducted the simulations, and analyzed the results.
S.~T. managed and supervised the research. 
M.~Y. and S.~T. contributed to the discussion on the conceptualization, methodology, results, and implications. 
All authors reviewed the manuscript.

\section*{Additional information}
\textbf{Competing interests}\\
The authors declare no competing interests.

\section*{Data Availability}
The datasets used in our study are available from the corresponding author upon reasonable request.

\end{document}